\documentclass[conference]{IEEEtran}
\IEEEoverridecommandlockouts

\usepackage[T1]{fontenc}
\usepackage[utf8]{inputenc}
\usepackage[english]{babel}
\usepackage{cite}
\usepackage{amsmath,amssymb,amsfonts}
\usepackage{graphicx}
\usepackage{textcomp}
\usepackage{xcolor}
\usepackage{subcaption}
\usepackage{booktabs}
\usepackage{url}
\usepackage[breaklinks=true,colorlinks=false,linkcolor=blue,urlcolor=blue,citecolor=blue]{hyperref}
\usepackage[subtle,tracking=normal]{savetrees}

\newcommand{\smalltt}[1]{{\small\texttt{#1}}}
\newcommand{\footnotett}[1]{{\footnotesize\texttt{#1}}}
\newcommand{\yahmp}{YAHMP}

\def\BibTeX{{\rm B\kern-.05em{\sc i\kern-.025em b}\kern-.08em
    T\kern-.1667em\lower.7ex\hbox{E}\kern-.125emX}}

\title{What Matters in Humanoid General Motion Tracking? An Empirical Study}

\author{
\IEEEauthorblockN{Fabio Amadio and Enrico Mingo Hoffman}
\thanks{The authors are with Inria, Universit\'{e} de Lorraine, CNRS, Nancy,
France. Emails: \{\footnotett{fabio.amadio},\footnotett{enrico.mingo-hoffman}\}\footnotett{@inria.fr}. This work was
supported by the ANR project MeRLin (ANR-24-CE33-0753-01).}
}

\begin{document}

\maketitle

\begin{abstract}
Humanoid general motion tracking requires policies that can follow diverse whole-body references while maintaining balance.
Building such policies involves many practical design choices, and their individual effects are often hard to assess.
We address this issue with an empirical study of common modeling and training factors used in recent humanoid motion-imitation pipelines.
To make the study controlled and reproducible, we developed \yahmp{}, an open-source modular framework for training, evaluating, and deploying whole-body motion tracking policies on the Unitree G1.
Within \yahmp{}, we define a nominal configuration and compare variants that differ in motion-command representation, observation history, action representation, actuation profile, hand-force randomization during training, and training approach.
We evaluate the resulting policies on a test set of retargeted human motions and compare the nominal policy with TWIST2 as an external baseline trained on the same motion set.
The results distinguish choices with clear tracking effects from choices that mainly change actuation effort, training complexity, or physical interaction capability.
Finally, we deploy \yahmp{} policies zero-shot on the real Unitree G1, demonstrating diverse whole-body motion tracking, balance under external perturbations, and forceful interaction.
\end{abstract}

\section{Introduction}
Example-guided reinforcement learning (RL) uses reference motions as a practical training signal for complex motor skills.
Instead of hand-designing a reward for each behavior, the policy is encouraged to reproduce a demonstrated motion while remaining physically plausible.
DeepMimic first showed this idea on simulated characters~\cite{peng2018deepmimic}, and later humanoid and legged-robot works used human or animal references to learn specific locomotion behaviors~\cite{peng2020learning,zhang2024humanreference}.
These results established motion imitation as a powerful tool for robot control, but imitating one reference motion, or a small set of related motions, is different from training a deployable controller that can follow a broad distribution of whole-body references.
\begin{figure}[t]
    \centering
    \includegraphics[width=\linewidth,trim={11.6cm 6.3cm 9.5cm 1.8cm},clip]{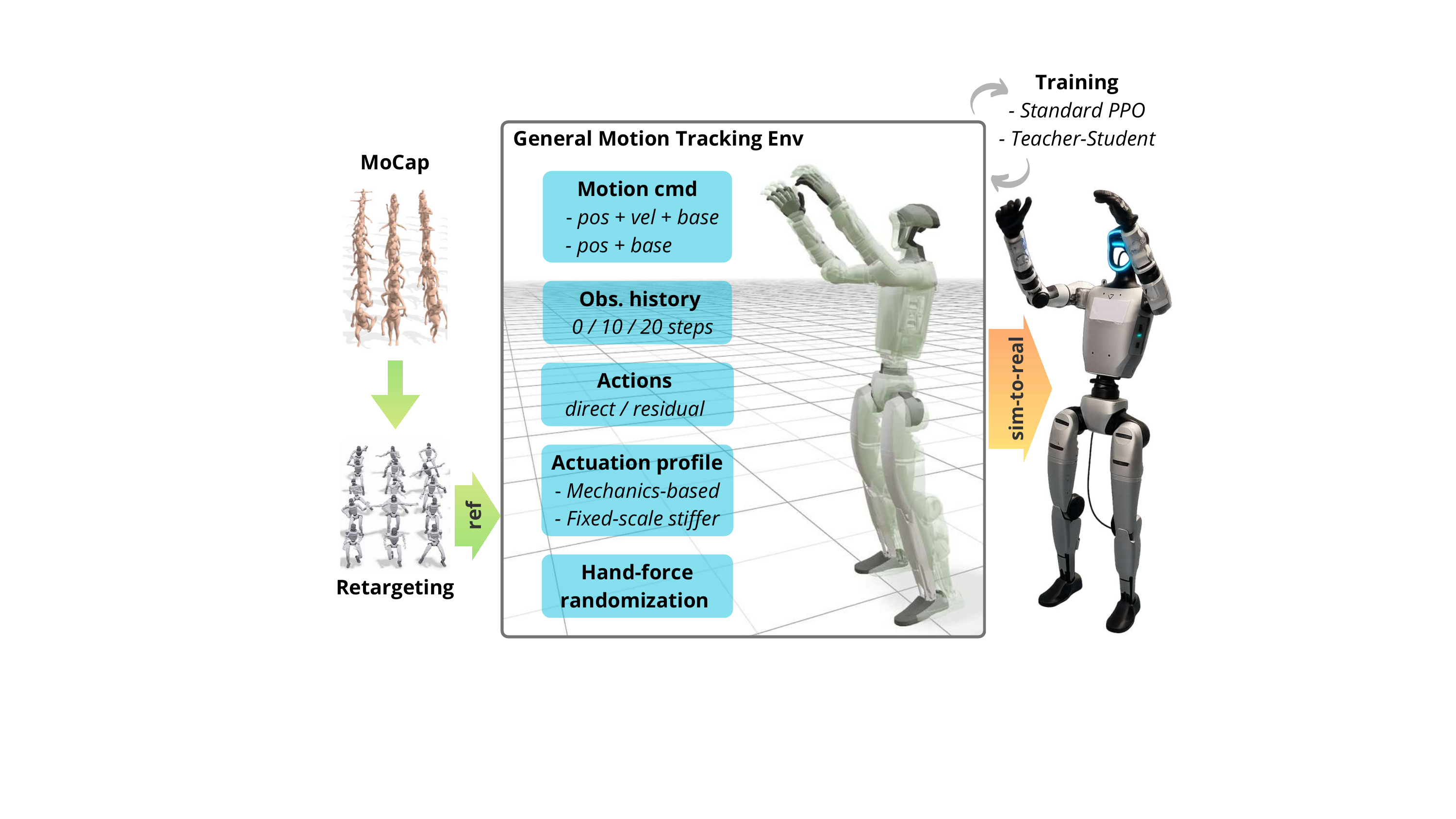}
    \caption{Overview of the \yahmp{} experimental pipeline: retargeted motions are used to train controlled policy variants, which are evaluated in simulation and on the Unitree G1.}
    \label{fig:yahmp_scheme}
\end{figure}
This setting is often referred to as \emph{general motion tracking} \cite{chen2025gmt}.
For humanoid robots, recent work uses this term for policies that track references spanning locomotion, turns, gestures, dancing, crouching, and whole-body interaction with objects and the environment~\cite{ji2024exbody2}, rather than specializing to a single clip.
In parallel, whole-body imitation has also been used as the tracking layer of teleoperation architectures~\cite{ze2025twist,ze2025twist2}.
Training such motion tracking systems relies on large-scale human motion and interaction datasets \cite{AMASS:ICCV:2019,li2023object}, together with retargeting methods that map human motion to the robot morphology~\cite{joao2025gmr}.

The current literature provides several complete pipelines, each with its own set of design choices.
However, it gives limited practical guidance on how individual factors affect general tracking performance.
Existing ablations are usually tied to a specific method, while comparisons between methods evaluate complete systems rather than individual design choices.
This makes it difficult to identify which settings are useful across implementations and which mainly reflect the details of a specific pipeline.

We therefore organize the study around factors that recur across recent whole-body tracking pipelines but are rarely analyzed systematically within the same controlled setup.
\textbf{(i)}~\textbf{Motion-command representation} controls which reference quantities are passed to the actor.
Recent systems commonly expose reference joint positions, base velocities, and key-body targets~\cite{ze2025twist,ji2024exbody2,chen2025gmt}, while explicit reference joint velocities are less standard.
\textbf{(ii)}~\textbf{Observation history} captures recent state and command evolution, which can help with partial observability and short-term dynamics~\cite{li2025reinforcement,ze2025twist2}.
\textbf{(iii)}~\textbf{Action representation} determines how policy outputs are converted into motor commands, either as offsets around a fixed default posture or as residual offsets around the reference joint positions~\cite{zhao2025resmimic}.
\textbf{(iv)}~\textbf{Actuation profile}, defined by PD gains and action scales, then determines how these joint targets are converted into torques~\cite{liao2025beyondmimic}.
\textbf{(v)}~\textbf{Hand-force randomization}: a policy can follow motions accurately while still failing to sustain meaningful interaction forces~\cite{zhang2025falcon}.
By applying random external forces to the hands during training, the policy is exposed to interaction conditions that are not captured by unperturbed free-space tracking.
\textbf{(vi)}~\textbf{Teacher-student training}: a policy can be trained directly with a single-stage PPO~\cite{schulman2017proximal}, or with a teacher-student approach in which a privileged teacher guides a deployable student policy~\cite{ze2025twist,he2024omnih2o,ji2024exbody2}.

To evaluate these factors under a common protocol, we developed \yahmp{} (\emph{Yet Another Humanoid Motion tracking Policy}), an open-source modular framework for training, evaluating, and deploying whole-body motion tracking policies on the Unitree G1, built on top of \texttt{mjlab}~\cite{zakka2026mjlab}.
Within \yahmp{}, we define a nominal configuration and instantiate controlled variants by changing one modeling or training component at a time.
Fig.~\ref{fig:yahmp_scheme} summarizes the experimental pipeline.

We evaluate the variants in simulation on test motions not used for training.
To put these results in context, we also retrain TWIST2 on the same training set and evaluate it with the same metrics.
We then deploy selected policies zero-shot on the real Unitree G1 to test dynamic motion tracking, tolerance to external perturbations, and meaningful interaction forces.

To summarize, the main contributions of this work are:
\begin{itemize}
    \item a controlled empirical study of modeling and training factors for humanoid general motion tracking;
    \item \yahmp{}, an open-source framework for training, evaluating, and deploying whole-body motion tracking policies for the Unitree G1 robot, with code publicly available at \href{https://github.com/hucebot/yahmp}{\smalltt{https://github.com/hucebot/yahmp}};
    \item a zero-shot sim-to-real evaluation of the nominal policy, together with hardware studies of hand-force randomization and actuation-profile effects.
\end{itemize}

The remainder of the paper is organized as follows.
Sec.~\ref{sec:framework} describes the \yahmp{} experimental framework and the design choices considered in the study.
Sec.~\ref{sec:simulation} reports the simulation evaluation, including the internal ablations and the comparison with TWIST2.
Sec.~\ref{sec:hardware} presents the real-robot deployment and the hardware experiments on hand-force randomization and actuation profile effects.
Finally, Sec.~\ref{sec:conclusion} discusses the results and draws conclusions.

\section{YAHMP Experimental Framework}
\label{sec:framework}
\yahmp{} is a configurable framework for training humanoid general motion tracking policies for the Unitree G1 robot.
It allows individual design choices to be varied within a common training and evaluation setup.
This section describes the nominal configuration first, and introduces the alternative settings that will be evaluated in Secs.~\ref{sec:simulation} and~\ref{sec:hardware}.

\subsection{General Motion Tracking Problem}
We formulate whole-body motion tracking as a goal-conditioned RL problem.
At time $t$, the policy receives an observation $\mathbf{o}_t$ containing the current robot state and a motion reference, and outputs an action $\mathbf{a}_t$ that is converted into joint-position targets for the low-level PD controllers.
Let $\mathbf{q}_t,\dot{\mathbf{q}}_t\in\mathbb{R}^n$ be the robot joint positions and velocities ($n=29$ in the case of the Unitree G1), and let the floating-base state include the base orientation, linear velocity, and angular velocity.
Reference motions are obtained by retargeting human motion-capture data to the G1 morphology~\cite{joao2025gmr}.
The policy is trained to maximize the expected return, with rewards that encourage tracking of the reference motion while regularizing contacts, actions, and joint-limit violations.

\subsection{Observations}

Within \yahmp{}, observations are built by combining a small set of configurable components.
The two components used by all deployable policies in this study are \emph{proprioception} and a \emph{motion command}.
A \emph{history buffer} can then be added to provide short-term context.
Privileged observations are available when required by the training pipeline.

The \emph{proprioception} vector contains deployable robot-state terms and the previous policy action.
\begin{equation}
\mathbf{p}_t = [{}^b\boldsymbol{\omega}_t,\ {}^b\mathbf{g}_t,\ \mathbf{q}_t-\mathbf{q}^{\mathrm{nom}},\ \dot{\mathbf{q}}_t,\ \mathbf{a}_{t-1}],
\end{equation}
where ${}^b\boldsymbol{\omega}_t$ is base angular velocity, ${}^b\mathbf{g}_t$ is the unit gravity direction expressed in the base frame, $\mathbf{q}^{\mathrm{nom}}$ is the default joint configuration, and $\mathbf{a}_{t-1}$ is the previous action.

The \emph{motion command} exposes a compact representation of the reference motion for the robot to track.
The nominal motion command is defined as
\begin{equation}
\mathbf{c}_t = [
\hat{\mathbf{q}}_t,\ \dot{\hat{\mathbf{q}}}_t,\ {}^b\hat{\mathbf{v}}_{xy,t},\
{}^b\hat \omega_{z,t},\ {}^w\hat h_t,\ {}^b\hat\phi_t,\ {}^b\hat\theta_t],
\end{equation}
where ${}^b\hat{\mathbf{v}}_{xy,t}$ is reference planar base velocity in the base frame, ${}^b\hat \omega_{z,t}$ is reference yaw rate, ${}^w\hat h_t$ is base height in the world frame, and ${}^b\hat\phi_t,{}^b\hat\theta_t$ are reference roll and pitch.
We also consider a \textbf{Pos-ref-only} variant that removes $\dot{\hat{\mathbf{q}}}_t$ from the command, keeping the other components unchanged.

The \emph{history buffer} concatenates past \emph{proprioception} and \emph{motion command} vectors:
\begin{equation}
\mathcal{H}_t = \{[\mathbf{c}_{t-H},\mathbf{p}_{t-H}], \ldots, [\mathbf{c}_{t-1},\mathbf{p}_{t-1}]\}.
\end{equation}
The nominal setting uses $H=10$ (\textbf{History-10}); we also test $H=20$ (\textbf{History-20}) and a \textbf{No history} variant.

Finally, \emph{privileged observations} include proprioceptive measurements without noise and additional simulation terms: base linear velocity, reference-relative base pose, robot body poses and orientations expressed in the world frame, foot contact state, and ground-friction coefficients.

\subsection{Actions}

Actions are interpreted as offsets applied to joint-position targets.
Inspired by ResMimic~\cite{zhao2025resmimic}, the nominal policy uses \textbf{Residual actions} around the reference joint configuration:
\begin{equation}
\mathbf{q}_t^{\mathrm{target}} = \hat{\mathbf{q}}_t + \boldsymbol{\alpha} \odot \mathbf{a}_t,
\end{equation}
where $\boldsymbol{\alpha}$ is a vector of joint-wise action scales and $\odot$ denotes element-wise multiplication.
We also test the \textbf{No residual} variant, where actions are instead applied around the fixed default posture:
\begin{equation}
\mathbf{q}_t^{\mathrm{target}} = \mathbf{q}^{\mathrm{nom}} + \boldsymbol{\alpha} \odot \mathbf{a}_t.
\end{equation}
PD controllers track joint targets with zero desired velocity:
\begin{equation}
\boldsymbol{\tau}_t =
\mathbf{K}_p(\mathbf{q}_t^{\mathrm{target}}-\mathbf{q}_t)
- \mathbf{K}_d\dot{\mathbf{q}}_t,
\end{equation}
with $\mathbf{K}_p$ and $\mathbf{K}_d$ diagonal positive-definite gain matrices.
Torques are saturated at the actuator effort limits.

\subsection{Actuation Profile}

We use the term \emph{actuation profile} to denote the combined choice of joint PD gains and action scales.
The PD gains determine how strongly the low-level controller tracks the joint targets, while the action scales map the policy outputs to joint-position offsets.

The nominal actuation profile is \textbf{Mechanics-based}.
Following BeyondMimic~\cite{liao2025beyondmimic}, the PD gains are selected from the reflected actuator inertia $I_j$:
\begin{equation}
k_{p,j}=I_j\omega_0^2,
\qquad
k_{d,j}=2\zeta I_j\omega_0,
\end{equation}
where $\omega_0=2\pi\cdot 10\,\mathrm{rad\,s^{-1}}$ and $\zeta=2$.
The joint-wise action scale is set from the actuator effort limit:
\begin{equation}
\alpha_j = 0.25\,\frac{\tau_{j,\max}}{k_{p,j}}.
\end{equation}
With this scaling, equal policy outputs correspond to equal fractions of the torque limit across joints.

We denote the alternative profile as \textbf{Stiffer fixed-scale}.
It follows settings used in recent humanoid motion tracking implementations~\cite{ze2025twist,ze2025twist2,zhang2025falcon}.
It replaces the mechanics-based gains with manually specified, generally higher PD gains and uses a constant action scale for all joints.
The numerical values used for both profiles are reported in Table~\ref{tab:appendix_actuation} of Appendix~\ref{app:implementation_details}.

\subsection{Actor-Critic Architecture}

The actor and critic networks use an MLP core with hidden dimensions $(512,512,256,128)$, running observation normalization, layer normalization in the last hidden layer, and ELU activations.
When a history buffer is used, it is encoded by a 1D temporal convolutional network, as commonly done in robust legged RL policies~\cite{li2025reinforcement}.
The history encoder has two temporal convolution layers with 48 and 24 channels, kernel sizes 6 and 4, strides 2 and 2, and a linear projection to a 64-dimensional history embedding.
The resulting history embedding is concatenated with the current proprioception and motion command before the MLP.
When privileged observations are enabled, they are provided only to training-time networks, such as the critic or the teacher policy.

\begin{table}[t]
\centering
\caption{Nominal configuration and considered variants.}
\label{tab:design_choices}
\scriptsize
\setlength{\tabcolsep}{3pt}
\begin{tabular}{lll}
\toprule
Design axis & Nominal & Alternatives \\
\midrule
Motion command & Pos. and vel. ref. & Pos-ref-only \\
Observation history & History-10 & No history; History-20 \\
Action representation & Residual actions & No residual \\
Actuation profile & Mechanics-based & Stiffer fixed-scale \\
Hand-force randomization & Disabled & Enabled \\
Policy training & PPO & Teacher-student \\
\bottomrule
\end{tabular}
\end{table}

\subsection{Domain Randomization}

The nominal domain randomization includes foot-ground friction, torso center of mass, uniform additive proprioceptive noise, and intermittent base-velocity pushes.

To study interaction capability, \yahmp{} also supports \textbf{Hand-force randomization}, a training setting with external forces applied at the hands.
This is inspired by the torque-limit-aware end-effector force sampling used in FALCON~\cite{zhang2025falcon}.
For each force event, \yahmp{} samples the side, application point, duration, and cooldown.
The application point is sampled around the wrist link to approximate contacts applied at different points on the hand.
The force is then scaled to remain feasible with respect to the joint-torque limits.
The randomization ranges are reported in Table~\ref{tab:appendix_dr} of Appendix~\ref{app:implementation_details}.

\subsection{Reward, Reset, and Motion References}

All variants use the same reward, which combines tracking terms for base pose, body pose, body velocity, joint position, and joint velocity with regularizers on foot contact forces, foot slip, action rate, and joint-limit violations.
The complete set of reward terms and coefficients is reported in Table~\ref{tab:appendix_reward} of Appendix~\ref{app:implementation_details}.

Episodes terminate when the reference motion expires or when selected bodies deviate too far from the reference.
At reset, \yahmp{} samples both a motion clip and an initial phase within the clip.
The sampling distribution is updated from tracking statistics, so poorly tracked clips and phases are replayed more often while a uniform component preserves coverage.

\begin{figure*}[t]
    \centering
    \includegraphics[width=\textwidth]{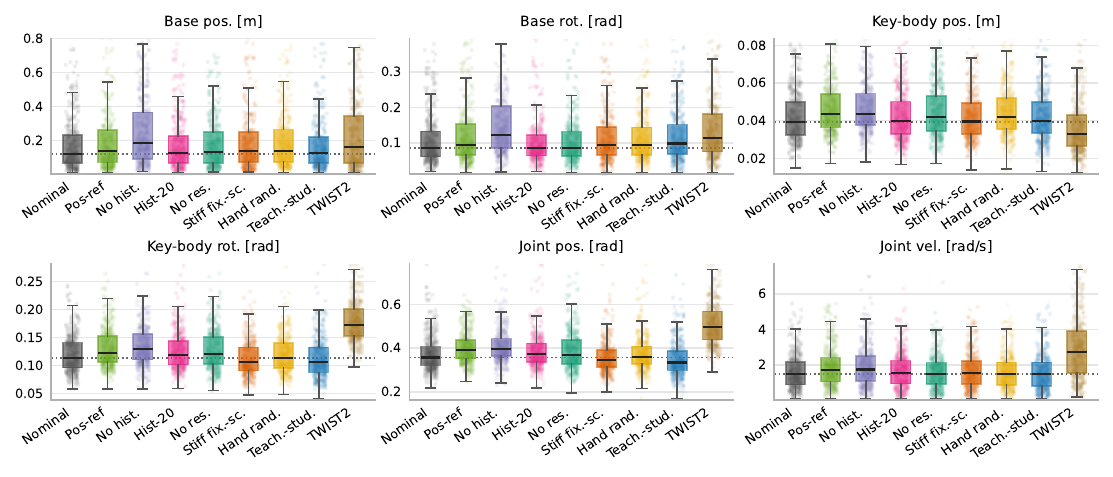}
    \caption{Tracking-error distributions over 1,024 test motions.
    Boxes show median, interquartile range, and 1.5-IQR whiskers.
    The dotted line marks the nominal policy median; translucent points show errors for a random subset of motions, while outliers are hidden.}
    \label{fig:tracking_ablation_boxplots}
\end{figure*}

\subsection{Policy Training and Dataset}

We train on retargeted AMASS~\cite{AMASS:ICCV:2019} and OMOMO~\cite{li2023object} motions.
After filtering unsuitable clips, the dataset contains 12,175 motions.
We use 11,151 motions for training and reserve 1,024 motions for testing.
The test set is selected from the filtered motion collection to cover a broad range of motion types and styles, including walking, dancing, gestures, stretching, object interaction, static poses, and quasi-static transitions.
The split was fixed before running the ablations and was not selected based on policy performance.
Train and test motions can originate from the same source datasets and subjects, but the retargeted motions used for evaluation are not used during training.

The nominal training pipeline uses PPO with an asymmetric actor-critic setup.
The actor receives only deployable observations: \emph{proprioception}, \emph{motion command}, and, when enabled, the \emph{history buffer}.
The critic can also receive \emph{privileged observations}.

\yahmp{} also includes a \textbf{Teacher-student} variant based on a teacher-guided RL pipeline~\cite{ze2025twist}.
In this case, we first train a teacher actor with PPO, using privileged information available only in simulation.
The teacher is then frozen, and the student receives only deployable observations and is trained with PPO while being regularized toward the teacher distribution by penalizing the Kullback--Leibler (KL) divergence:
\begin{equation}
\mathcal{L}_{\mathrm{student}}
=
\mathcal{L}_{\mathrm{PPO}}
\;+\;
\lambda\,D_{\mathrm{KL}}\!\left(
\pi_{\mathrm{student}}
\;\|\;
\pi_{\mathrm{teacher}}
\right).
\end{equation}
Both teacher and student are trained for 20,000 iterations, with the KL coefficient $\lambda$ linearly annealed from 0.1 to 0.07 during student training.

All \yahmp{} variants are trained with 8,192 parallel environments for 20,000 PPO iterations.
Training runs use a single NVIDIA GeForce RTX 4090, and each run required approximately 25 hours to complete.
The PPO hyperparameters are reported in Table~\ref{tab:appendix_training} of Appendix~\ref{app:implementation_details}.

Table~\ref{tab:design_choices} summarizes the characteristics of the nominal configuration and the variants considered in the study.

\section{Simulation Evaluation}
\label{sec:simulation}
In this section, we evaluate the different \yahmp{} variants in simulation and compare the nominal policy with TWIST2~\cite{ze2025twist2}.
All policies use the same Unitree G1 model and are evaluated on the same 1,024 test motions in MuJoCo~\cite{todorov2012mujoco}, with a simulation time step of 0.005\,s and a control frequency of 50\,Hz.
Each rollout starts from the first frame of the reference motion and runs without external perturbations.
A rollout is successful if the robot completes the motion without falling, defined as contact between the torso and the ground.
All policies complete the full test set without falling, so the evaluation focuses on tracking accuracy.

We measure the base-position error ($e_{\mathrm{base}}^p$), base-orientation error ($e_{\mathrm{base}}^R$), mean key-body position and orientation errors ($e_{\mathrm{bodies}}^p$ and $e_{\mathrm{bodies}}^R$) over nine bodies (torso, knees, ankles, elbows, and wrists), and joint-position and joint-velocity errors ($e_q$ and $e_{\dot q}$).
For each policy, tracking errors are first averaged over time for each motion and then aggregated over the 1,024 test motions.
Fig.~\ref{fig:tracking_ablation_boxplots} shows their distributions, while Table~\ref{tab:ablation_results} reports the corresponding means, standard deviations, and relative changes with respect to the nominal policy.

\begin{table*}[t]
\centering
\caption{Tracking performance over 1,024 test motions.
Entries report mean $\pm$ standard deviation across motions; parentheses show the mean change relative to the nominal policy.
Positive changes mean larger error.
Bold marks the lowest mean error per metric.
All policies achieve 100\% success; TWIST2 is retrained on the same training motion set.
}
\label{tab:ablation_results}
\scriptsize
\setlength{\tabcolsep}{1.7pt}
\begin{tabular}{l r@{$\pm$}l r@{$\pm$}l r@{$\pm$}l r@{$\pm$}l r@{$\pm$}l r@{$\pm$}l}
\toprule
Variant & \multicolumn{2}{c}{$e_{\mathrm{base}}^p$ [m]} & \multicolumn{2}{c}{$e_{\mathrm{base}}^R$ [rad]} & \multicolumn{2}{c}{$e_{\mathrm{bodies}}^p$ [m]} & \multicolumn{2}{c}{$e_{\mathrm{bodies}}^R$ [rad]} & \multicolumn{2}{c}{$e_q$ [rad]} & \multicolumn{2}{c}{$e_{\dot q}$ [rad/s]} \\
\midrule
Nominal & $0.197$ & $0.233$ & $0.121$ & $0.125$ & $0.044$ & $0.020$ & $0.123$ & $0.047$ & $0.377$ & $0.093$ & $1.69$ & $1.11$ \\
\midrule
Pos-ref-only & $0.211$ & $0.248$ {\scriptsize(+7\%)} & $0.138$ & $0.158$ {\scriptsize(+14\%)} & $0.048$ & $0.021$ {\scriptsize(+9\%)} & $0.134$ & $0.052$ {\scriptsize(+9\%)} & $0.408$ & $0.090$ {\scriptsize(+8\%)} & $1.95$ & $1.31$ {\scriptsize(+15\%)} \\
No history & $0.273$ & $0.264$ {\scriptsize(+38\%)} & $0.181$ & $0.192$ {\scriptsize(+50\%)} & $0.048$ & $0.017$ {\scriptsize(+10\%)} & $0.136$ & $0.042$ {\scriptsize(+11\%)} & $0.415$ & $0.089$ {\scriptsize(+10\%)} & $1.98$ & $1.25$ {\scriptsize(+17\%)} \\
History-20 & $0.186$ & $0.195$ {\scriptsize(-6\%)} & $\mathbf{0.112}$ & $\mathbf{0.102}$ {\scriptsize\textbf{(-7\%)}} & $0.044$ & $0.020$ {\scriptsize(+1\%)} & $0.128$ & $0.051$ {\scriptsize(+4\%)} & $0.393$ & $0.094$ {\scriptsize(+4\%)} & $1.75$ & $1.15$ {\scriptsize(+4\%)} \\
No residual & $0.199$ & $0.212$ {\scriptsize(+1\%)} & $0.115$ & $0.101$ {\scriptsize(-5\%)} & $0.046$ & $0.018$ {\scriptsize(+5\%)} & $0.130$ & $0.045$ {\scriptsize(+6\%)} & $0.395$ & $0.108$ {\scriptsize(+5\%)} & $1.69$ & $1.07$ {\scriptsize(0\%)} \\
Stiffer fixed-scale & $0.204$ & $0.227$ {\scriptsize(+4\%)} & $0.130$ & $0.139$ {\scriptsize(+8\%)} & $0.043$ & $0.015$ {\scriptsize(-2\%)} & $0.114$ & $0.035$ {\scriptsize(-7\%)} & $0.364$ & $0.084$ {\scriptsize(-3\%)} & $1.73$ & $1.14$ {\scriptsize(+2\%)} \\
Hand-force rand. & $0.204$ & $0.201$ {\scriptsize(+4\%)} & $0.125$ & $0.110$ {\scriptsize(+4\%)} & $0.046$ & $0.018$ {\scriptsize(+5\%)} & $0.122$ & $0.043$ {\scriptsize(-1\%)} & $0.379$ & $0.090$ {\scriptsize(+1\%)} & $1.68$ & $1.08$ {\scriptsize(-1\%)} \\
Teacher-student & $\mathbf{0.181}$ & $\mathbf{0.196}$ {\scriptsize\textbf{(-8\%)}} & $0.130$ & $0.112$ {\scriptsize(+8\%)} & $0.044$ & $0.015$ {\scriptsize(0\%)} & $\mathbf{0.113}$ & $\mathbf{0.036}$ {\scriptsize\textbf{(-8\%)}} & $\mathbf{0.353}$ & $\mathbf{0.091}$ {\scriptsize\textbf{(-6\%)}} & $\mathbf{1.65}$ & $\mathbf{1.10}$ {\scriptsize\textbf{(-3\%)}} \\
\midrule
TWIST2 & $0.260$ & $0.289$ {\scriptsize(+32\%)} & $0.165$ & $0.184$ {\scriptsize(+36\%)} & $\mathbf{0.039}$ & $\mathbf{0.031}$ {\scriptsize\textbf{(-10\%)}} & $0.186$ & $0.083$ {\scriptsize(+51\%)} & $0.521$ & $0.142$ {\scriptsize(+38\%)} & $3.06$ & $2.12$ {\scriptsize(+81\%)} \\
\bottomrule
\end{tabular}
\end{table*}

\begin{table*}[t]
\centering
\caption{Torque comparison for the actuation-profile variants over 1,024 test motions.
Entries report mean $\pm$ standard deviation in N\,m across motions; maximum torques are first computed over each rollout and then aggregated across motions.
}
\label{tab:actuation_torque}
\scriptsize
\setlength{\tabcolsep}{3pt}
\begin{tabular}{lcccccc}
\toprule
Actuation profile & Mean all & Max all & Mean upper & Max upper & Mean lower & Max lower \\
\midrule
Mechanics-based & $\mathbf{3.56 \pm 0.89}$ & $\mathbf{51.2 \pm 21.0}$ & $\mathbf{1.52 \pm 0.51}$ & $\mathbf{22.8 \pm 11.1}$ & $\mathbf{6.46 \pm 1.61}$ & $\mathbf{51.2 \pm 21.0}$ \\
Stiffer fixed-scale & $3.63 \pm 0.94$ {\scriptsize(+2\%)} & $57.7 \pm 24.4$ {\scriptsize(+13\%)} & $1.58 \pm 0.54$ {\scriptsize(+4\%)} & $32.2 \pm 17.3$ {\scriptsize(+41\%)} & $6.53 \pm 1.69$ {\scriptsize(+1\%)} & $57.4 \pm 24.5$ {\scriptsize(+12\%)} \\
\bottomrule
\end{tabular}
\end{table*}

\subsection{Ablation Study}

For the \yahmp{} ablations, we change one design choice at a time from the nominal configuration.
We also repeated the ablation study with an additional training seed and obtained trends consistent with those reported below.

\subsubsection{Motion-Command Representation}

Removing the reference joint velocity from the motion command degrades all reported tracking metrics.
With the Pos-ref-only command, base errors increase by 7--14\%, key-body errors by about 9\%, joint-position error by 8\%, and joint-velocity error by 15\%.
Fig.~\ref{fig:tracking_ablation_boxplots} shows the same effect beyond the mean values, especially for base orientation and joint-velocity tracking.
The degradation remains visible even though the policy still receives observation history.
This suggests that the history encoder does not fully recover the phase and velocity information carried by $\dot{\hat{\mathbf{q}}}_t$.
These results indicate that including reference joint velocities alongside reference joint positions improves tracking performance.

\subsubsection{Observation History}

Without history, all tracking errors increase with respect to the nominal policy: base errors rise by 38--50\%, key-body errors by 10--11\%, joint-position error by 10\%, and joint-velocity error by 17\%.
Fig.~\ref{fig:tracking_ablation_boxplots} shows the same trend at the distribution level, with the No history distributions shifted upward across all reported metrics.
This supports the importance of providing the policy with temporal context.
Increasing the history length from 10 to 20 steps does not give a consistent improvement.
History-20 reduces the two base errors by 6--7\%, but slightly worsens key-body and joint-space tracking.
In particular, key-body orientation, joint-position, and joint-velocity errors increase by about 4\%.
Hence, History-20 does not provide a clear advantage over History-10, which provides a good balance between temporal context and input compactness.

\subsubsection{Action Representation}

When actions are applied around the fixed default posture instead of the reference joint positions (No residual), key-body pose errors increase by 5--6\%, and joint-position error increases by 5\%.
The effect on base tracking is less clear: base-position error increases by 1\%, while base-orientation error decreases by 5\% for the non-residual policy.
Overall, residual actions provide a modest improvement in key-body and joint-position tracking, while their effect on base tracking remains mixed.

\subsubsection{Actuation Profile}

For this ablation, tracking errors alone do not capture the relevant trade-off.
We therefore also report torque metrics in Table~\ref{tab:actuation_torque}, using mean and maximum absolute joint torques over all joints and over the upper and lower body separately.
Stiffer fixed-scale does not provide a clear tracking improvement.
It reduces key-body position and orientation errors and joint-position error, but increases both base errors and joint-velocity error.
The torque metrics show a clearer trend: Stiffer fixed-scale increases maximum torque by 13\% over all joints and by 41\% on the upper body.
Mean torques change less, but they also increase in all reported groups.
Thus the mechanics-based profile achieves comparable tracking performance while avoiding the larger torque peaks of Stiffer fixed-scale.

\subsubsection{Hand-Force Randomization}

In unperturbed simulation, hand-force randomization only mildly affects tracking, so we analyze its main effect on the real robot in Sec.~\ref{sec:hardware}.

\subsubsection{Teacher-Student Training}

Compared with the nominal PPO policy, Teacher-student reduces base-position and key-body orientation errors by 8\%, joint-position error by 6\%, and joint-velocity error by 3\%.
Key-body position error remains unchanged, while base-orientation error increases by 8\%.
Overall, these mixed changes amount to only minor improvements despite the additional training complexity.

\subsection{Comparison with TWIST2}

We use TWIST2 as an external complete-pipeline baseline.
Although it is presented as part of a teleoperation stack, its low-level controller addresses the same general motion tracking problem considered here.
We retrain it with its original code on the same 11,151 training motions and for the same number of PPO iterations used for \yahmp{}.
Both pipelines optimize base, body, and joint tracking, allowing the metrics defined above to provide a common evaluation of task-level tracking performance.

TWIST2 obtains the lowest key-body position error, reducing it by 10\% with respect to the nominal policy.
The remaining tracking errors are higher: base errors increase by 32--36\%, key-body orientation error by 51\%, joint-position error by 38\%, and joint-velocity error by 81\%.
TWIST2 therefore tracks key-body positions accurately on this test set, but shows poorer orientation and joint-space tracking under the common evaluation protocol.

\section{Real-Robot Evaluation}
\label{sec:hardware}

We use the real-robot experiments to test whether policies trained in simulation can be deployed on the Unitree G1 and to study the design choices whose effects are most relevant on hardware.
We first deploy the nominal policy for general motion tracking, then evaluate hand-force randomization under external loads and the effects of different actuation profiles.
The supplementary video, available at \href{https://youtu.be/BH6FpQzwm8M}{\smalltt{{https://youtu.be/BH6FpQzwm8M}}}, shows all real-robot experiments reported in this section.

Policy inference runs on an Intel Core Ultra 7 165H laptop with a direct cable connection to the robot.
At each step, the robot state is read through ROS~2 and combined with the reference motion; the resulting observation is passed to the policy, and joint-position targets are sent back to the robot within a 50\,Hz control loop (the same frequency used in simulation).

\subsection{Sim-to-Real Deployment}

\begin{figure}[t]
    \centering
    \begin{subfigure}{0.95\linewidth}
        \centering
        \includegraphics[width=\linewidth]{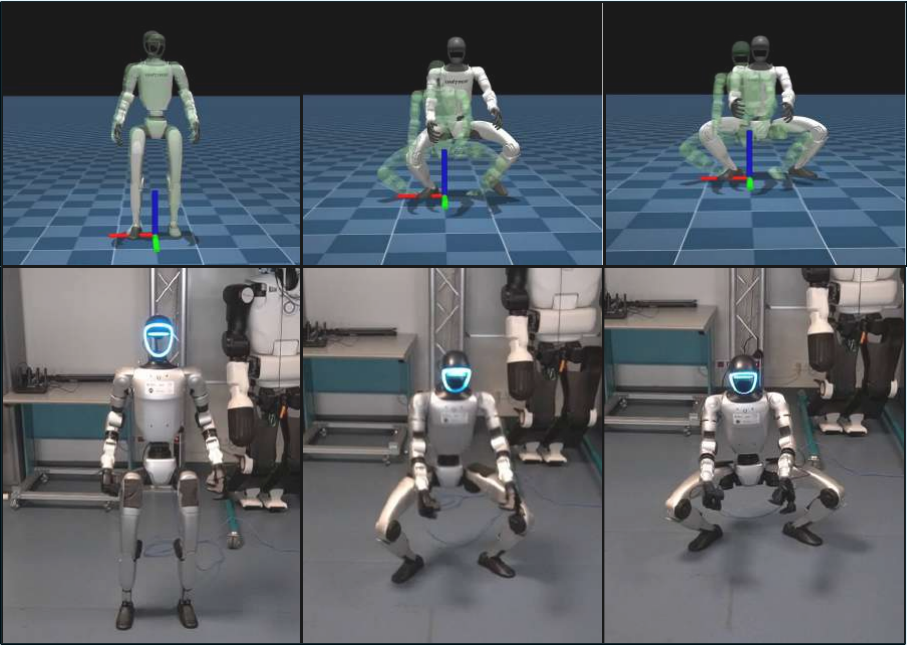}
        \caption{Squat-like motion.}
    \end{subfigure}
    \begin{subfigure}{\linewidth}
        \centering
        \includegraphics[width=0.95\linewidth]{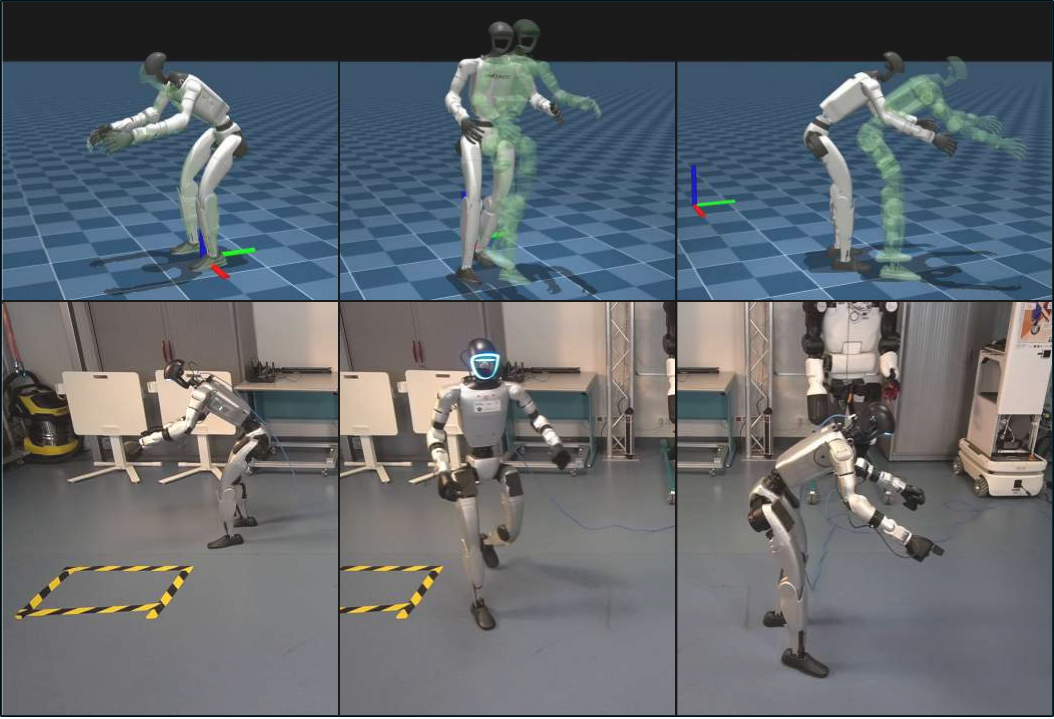}
        \caption{Loco-manipulation-style motion.}
    \end{subfigure}
    \caption{Motion tracking examples in simulation and on the Unitree G1.
    The green ghost shows the reference motion.}
    \label{fig:robot_examples}
\end{figure}

We first deploy the nominal policy on the Unitree G1 and use it to track different whole-body motions.
The same checkpoint trained in simulation runs directly on the robot, without real-robot fine-tuning, a dedicated sim-to-real pipeline, or additional filtering/smoothing of the policy commands.
The robot tracks references spanning locomotion, crouching, dancing, and loco-manipulation-style motions.
Motions can also be repeated or concatenated without resetting the controller.
Fig.~\ref{fig:robot_examples} shows representative examples.

\begin{figure}[h]
    \centering
    \includegraphics[width=0.19\linewidth,trim={10.6cm 0.85cm 13.1cm 1.69cm},clip]{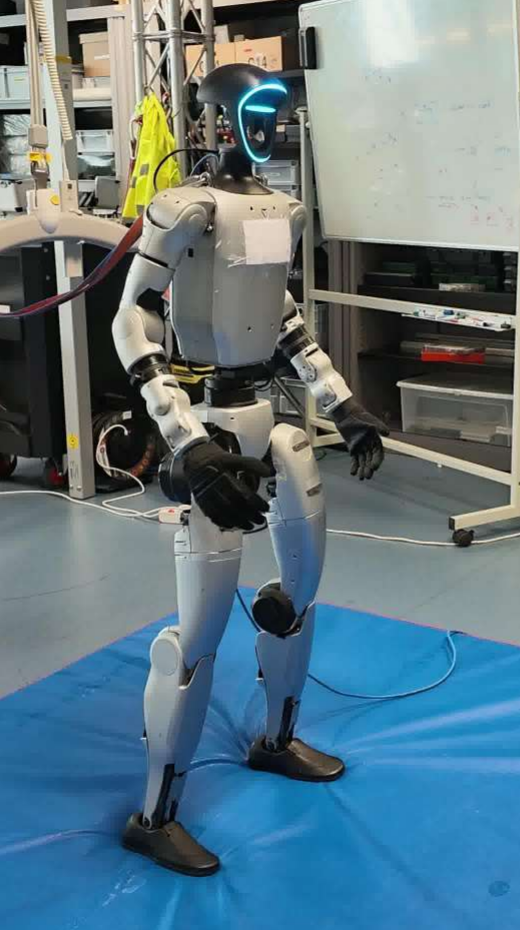}
    \includegraphics[width=0.19\linewidth,trim={10.6cm 0.85cm 13.1cm 1.69cm},clip]{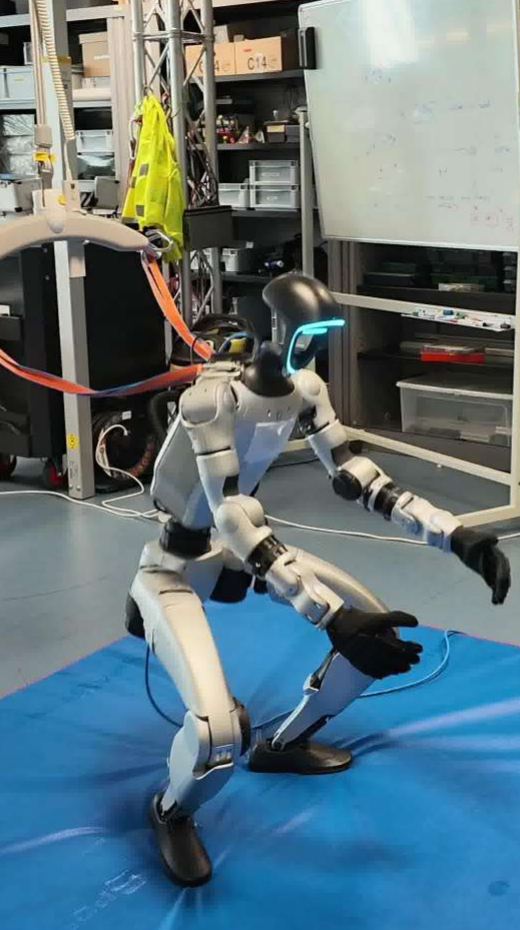}
    \includegraphics[width=0.19\linewidth,trim={10.6cm 0.85cm 13.1cm 1.69cm},clip]{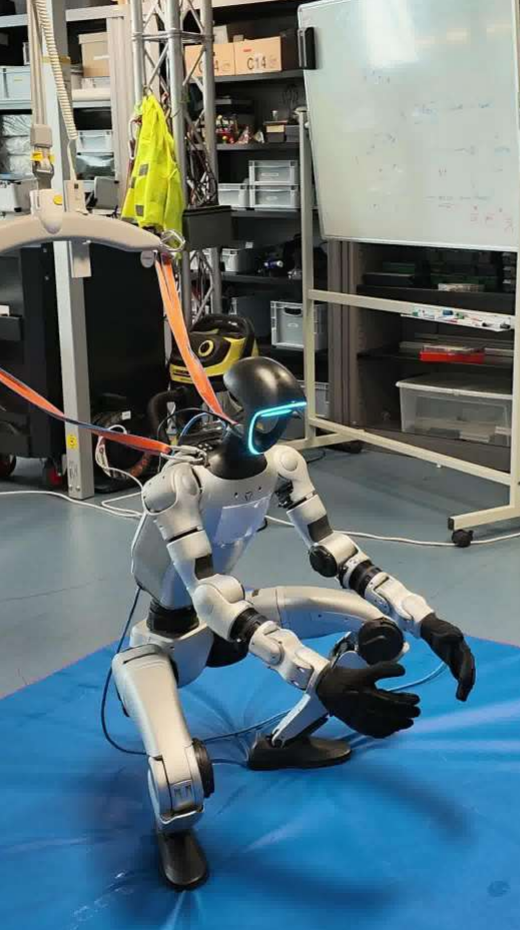}
    \includegraphics[width=0.19\linewidth,trim={10.6cm 0.85cm 13.1cm 1.69cm},clip]{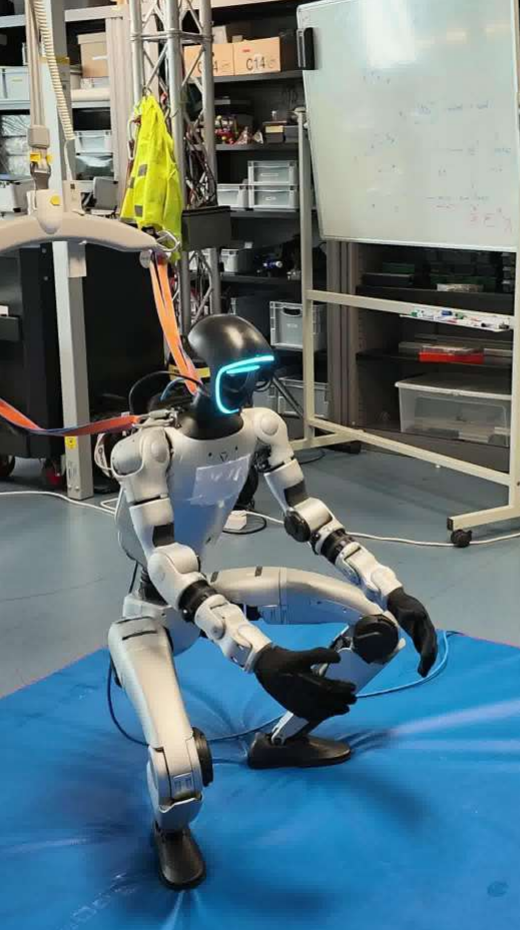}
    \includegraphics[width=0.19\linewidth,trim={10.6cm 0.85cm 13.1cm 1.69cm},clip]{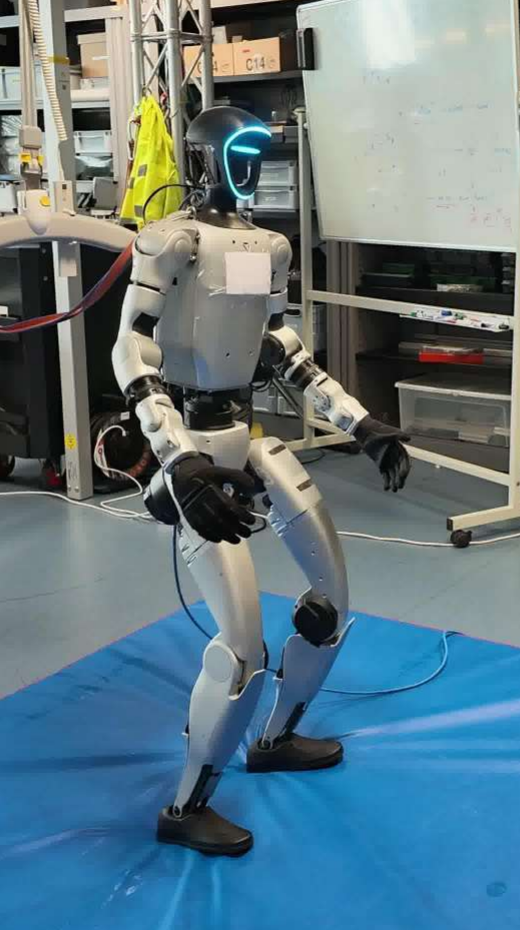}
    \caption{Execution of the squat motion on a soft mattress by the nominal policy without terrain-specific training.}
    \label{fig:mattress_squat}
\end{figure}

Beyond motion tracking, balance and robustness are also important for real-robot deployment.
We tested these aspects empirically by manually pushing and dragging the robot while it replayed reference motions.
The robot maintained or recovered its balance without resetting the policy.
We also tested how the controller reacts to unseen environmental conditions.
We placed the robot on a soft mattress, a ground condition never seen during training.
The robot was still able to track double-support motions, such as crouches and manipulation-like motions, while keeping balance (Fig.~\ref{fig:mattress_squat}), although dedicated training would still be necessary to achieve stable locomotion.

\subsection{Forceful Interactions via Hand-Force Randomization}

\begin{figure}[h]
    \centering
    \begin{subfigure}{0.23\linewidth}
        \centering
        \includegraphics[width=0.95\linewidth,trim={0cm 0.5cm 0cm 0.5cm},clip]{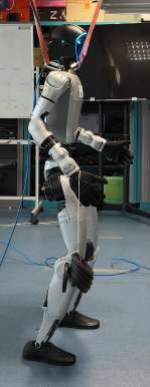}
        \caption{No force rand., 3\,kg.}
    \end{subfigure}
    \begin{subfigure}{0.23\linewidth}
        \centering
        \includegraphics[width=0.95\linewidth,trim={0cm 0.5cm 0cm 0.5cm},clip]{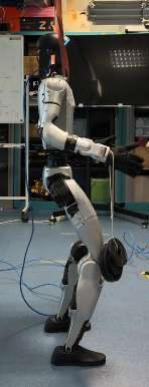}
        \caption{Hand-force rand., 3\,kg.}
    \end{subfigure}
    \begin{subfigure}{0.23\linewidth}
        \centering
        \includegraphics[width=0.95\linewidth,trim={0cm 0.5cm 0cm 0.5cm},clip]{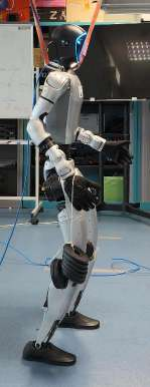}
        \caption{No force rand., 4\,kg.}
    \end{subfigure}
    \begin{subfigure}{0.23\linewidth}
        \centering
        \includegraphics[width=0.95\linewidth,trim={0cm 0.5cm 0cm 0.5cm},clip]{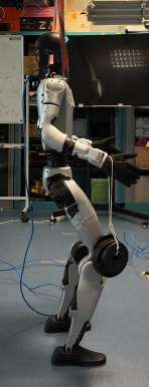}
        \caption{Hand-force rand., 4\,kg.}
    \end{subfigure}
    \caption{Representative frames from the hand-load experiment.
    Training with hand-force randomization improves the arm's ability to sustain external loads.}
    \label{fig:hand_load}
\end{figure}

Good free-space tracking does not necessarily imply that a policy can sustain an external load with its hands.
We study this distinction using two otherwise matched policies, trained with and without hand-force randomization.
On the robot, the right arm tracks the same fixed configuration for both policies while loads are progressively attached to the hand.
For each policy, loads were increased until the arm no longer reached a steady hold.
We measure the right-elbow deviation from the unloaded stable position at each load.
Table~\ref{tab:hand_loads} reports one controlled execution per policy, using the mean displacement over the steady window after each load is attached, while Fig.~\ref{fig:hand_load} shows representative configurations at 3 and 4\,kg.
The policy trained with hand-force randomization maintains the commanded posture under larger loads, whereas the policy trained without it exhibits substantially larger displacement.
At the largest load tested for both policies, 4\,kg, hand-force randomization reduces the elbow displacement from 15.5$^\circ$ to 6.6$^\circ$.
The hand-force-randomized policy also sustained loads of 5 and 6\,kg.
In unperturbed simulation, the same training setting keeps tracking close to the nominal policy, with only small changes across the reported metrics.
This indicates that hand-force randomization mainly affects the ability to exert meaningful forces with the hands, rather than nominal free-space tracking.

\begin{table}[b]
\centering
\caption{Right-elbow displacement from the unloaded stable posture during the hand-load experiment.
Values are means over steady hold windows, in degrees.}
\label{tab:hand_loads}
\small
\setlength{\tabcolsep}{3.3pt}
\begin{tabular}{lcccccc}
\toprule
Hand-force rand. & 1\,kg & 2\,kg & 3\,kg & 4\,kg & 5\,kg & 6\,kg \\
\midrule
Disabled & 3.2$^\circ$ & 8.3$^\circ$ & 13.3$^\circ$ & 15.5$^\circ$ & -- & -- \\
Enabled & 1.5$^\circ$ & 3.0$^\circ$ & 3.0$^\circ$ & 6.6$^\circ$ & 7.9$^\circ$ & 9.4$^\circ$ \\
\bottomrule
\end{tabular}
\vspace{-0.5cm}
\end{table}

\begin{figure}[h]
    \centering
    \includegraphics[width=0.307\linewidth,trim={3.39cm 0.42cm 8.47cm 1.69cm},clip]{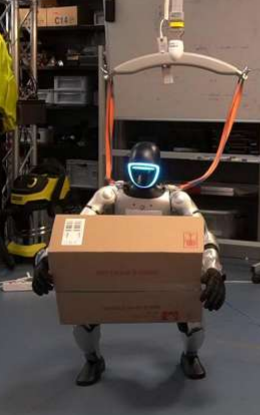}
    \includegraphics[width=0.307\linewidth,trim={3.39cm 0.42cm 8.47cm 1.69cm},clip]{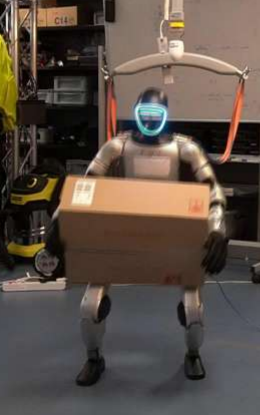}
    \includegraphics[width=0.307\linewidth,trim={3.39cm 0.42cm 8.47cm 1.69cm},clip]{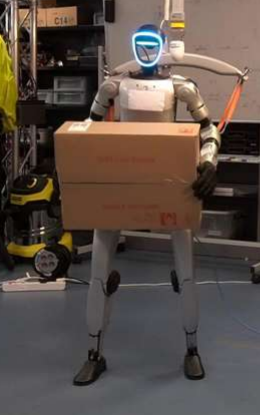}
    \caption{Frames from the full-body box-lifting replay.}
    \label{fig:box_lifting}
\end{figure}

As a qualitative demonstration, we use the hand-force-randomized policy to replay a full-body lifting motion (Fig.~\ref{fig:box_lifting}).
The robot starts crouched with a 1.5\,kg box held between its hands and stands while lifting the box.
This is not autonomous manipulation: the box state is not observed, and the reference is neither adapted nor replanned.
The experiment instead illustrates how the added force capacity can support a simple whole-body manipulation replay.

\subsection{Actuation Profile Effects in Sim-to-Real}

\begin{figure}[h]
    \centering
    \includegraphics[width=0.90\linewidth]{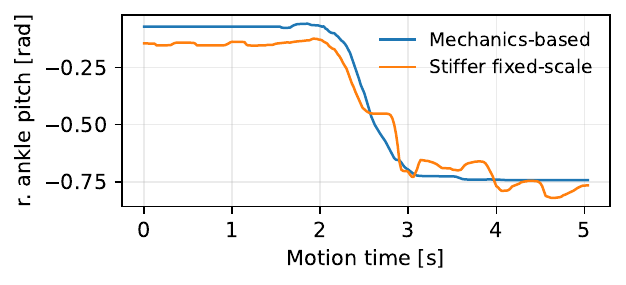}
    \caption{Measured right ankle pitch angle during a crouching deployment.
    The Stiffer fixed-scale policy exhibits larger oscillations than the Mechanics-based policy after the transition.}
    \label{fig:stiffpd_crouch_ankle_pitch}
\end{figure}

Finally, we deploy the Stiffer fixed-scale policy to determine whether the actuation trends observed in simulation persist on hardware.
The policy deploys successfully and completes the tested motions, but the crouching example shows a small increase in oscillation compared with the Mechanics-based policy, particularly at the ankles.
Fig.~\ref{fig:stiffpd_crouch_ankle_pitch} shows the right ankle pitch angle as a representative example from a matched crouching motion.
The Stiffer fixed-scale policy tracks the same overall motion but introduces visibly larger oscillations after the main crouch transition.
Smaller oscillations are also visible in other ankle and waist joints in this deployment.
Over the same motion phase, the mean absolute torque over all joints remains almost unchanged (4.52\,N\,m for the Mechanics-based policy and 4.53\,N\,m for Stiffer fixed-scale), but the maximum absolute torque increases from 54.5\,N\,m to 75.2\,N\,m.
On the plotted right ankle pitch joint, the mean absolute torque increases from 7.60\,N\,m to 10.38\,N\,m.
The main effects of the stiffer profile are larger torque peaks and more oscillatory behavior, while the average torque over all joints changes little.

\section{Conclusion}
\label{sec:conclusion}

In this paper, we presented an empirical analysis of common design choices in humanoid general motion tracking pipelines.
To make the comparison controlled and reproducible, we introduced \yahmp{}, an open-source framework for training, evaluating, and deploying whole-body motion tracking policies on the Unitree G1.
We used this framework to evaluate different variants of the same motion tracking pipeline, both in simulation and on the real robot.

The results show that including reference joint velocities in the motion command improves tracking performance, although this information is not always included explicitly in recent motion tracking pipelines.
A short history buffer also provides useful temporal information to the controller and likely helps the policy infer recent motion trends and compensate for partial observability in the deployable state.
The action-representation ablation is less decisive: residual actions provide a modest improvement in key-body and joint-position tracking, but no consistent advantage in base tracking.
The mechanics-based selection of PD gains and action scales also helps limit torque peaks while still tracking the desired motions accurately.
Regarding the training approach, standard PPO with deployable observations from the start already provides a strong nominal policy in our setup, while the more involved teacher-student pipeline gives only minor gains despite the added training complexity.

The real-robot experiments show that \yahmp{} policies transfer zero-shot to the Unitree G1, track diverse whole-body motions, and maintain balance under external perturbations.
They also show that hand-force randomization during training is important for enabling the policy to exert meaningful forces, rather than only tracking motions in free space.

The study focuses on one robot platform and varies one design choice at a time around a single nominal configuration, leaving interactions between choices and their effects on other humanoid platforms outside the current evaluation.

Future work will focus on integrating \yahmp{} with teleoperation and learning-from-demonstration systems, using the policy as the low-level whole-body tracking layer.
Teleoperation can provide structured demonstrations for coordinated manipulation skills~\cite{amadio2022target}; the resulting skills can then be integrated into autonomous robot stacks driven by user instructions~\cite{amadio2024vocal}.

\bibliographystyle{IEEEtran}
\bibliography{references}

\appendices
\section{Implementation Details}
\label{app:implementation_details}
The following tables summarize the PPO hyperparameters, actuation profiles, domain-randomization ranges, and reward terms used in the experiments.
Unless indicated otherwise, these settings are shared across all \yahmp{} variants.

\begin{table}[!h]
\centering
\caption{PPO hyperparameters used in \yahmp{}.}
\label{tab:appendix_training}
\scriptsize
\setlength{\tabcolsep}{2pt}
\begin{tabular}{@{}p{0.38\columnwidth}cc@{}}
\toprule
Parameter & PPO & Teacher-student \\
\midrule
Rollout steps/env & 24 & 24 \\
Learning epochs & 5 & 5 \\
Mini-batches & 4 & 4 \\
Learning rate & $10^{-3}$ & $3\cdot10^{-4}$ \\
LR schedule & adaptive & adaptive \\
Discount $\gamma$ & 0.99 & 0.99 \\
GAE $\lambda$ & 0.95 & 0.95 \\
Clip parameter & 0.2 & 0.2 \\
Entropy coefficient & 0.005 & 0.005 \\
Value-loss coefficient & 1.0 & 1.0 \\
Desired KL & 0.01 & 0.008 \\
Max gradient norm & 1.0 & 1.0 \\
Initial policy std. & 1.0 & 0.4 \\
Teacher KL coefficient & -- & $0.1\rightarrow0.07$ \\
\bottomrule
\end{tabular}
\end{table}

\begin{table}[!h]
\centering
\caption{PD gains and action scales considered.
Values are shared across left and right sides when applicable.}
\label{tab:appendix_actuation}
\scriptsize
\resizebox{\columnwidth}{!}{%
\begin{tabular}{lccc ccc}
\toprule
& \multicolumn{3}{c}{Mechanics-based} & \multicolumn{3}{c}{Stiffer fixed-scale} \\
Joint group & $k_p$ & $k_d$ & $\alpha$ & $k_p$ & $k_d$ & $\alpha$ \\
\midrule
Hip pitch/yaw & 40.18 & 2.56 & 0.548 & 100 & 2.0 & 0.5 \\
Hip roll & 99.10 & 6.31 & 0.351 & 100 & 2.0 & 0.5 \\
Knee & 99.10 & 6.31 & 0.351 & 150 & 4.0 & 0.5 \\
Ankle pitch/roll & 28.50 & 1.81 & 0.439 & 40 & 2.0 & 0.5 \\
Waist yaw & 40.18 & 2.56 & 0.548 & 150 & 4.0 & 0.5 \\
Waist roll/pitch & 28.50 & 1.81 & 0.439 & 150 & 4.0 & 0.5 \\
Shoulder/elbow & 14.25 & 0.91 & 0.439 & 40 & 5.0 & 0.5 \\
Wrist roll & 14.25 & 0.91 & 0.439 & 4 & 0.2 & 0.5 \\
Wrist pitch/yaw & 16.78 & 1.07 & 0.075 & 4 & 0.2 & 0.5 \\
\bottomrule
\end{tabular}
}
\end{table}

\begin{table}[!h]
\centering
\caption{Domain-randomization ranges used during training.}
\label{tab:appendix_dr}
\scriptsize
\setlength{\tabcolsep}{3pt}
\begin{tabular}{@{}p{0.38\columnwidth}p{0.55\columnwidth}@{}}
\toprule
Quantity & Range \\
\midrule
Foot-ground friction & $[0.3,1.2]$ \\
Torso COM offset & $x:\pm0.025$\,m, $y,z:\pm0.05$\,m \\
Base angular velocity noise & $\pm0.2$\,rad/s \\
Projected gravity noise & $\pm0.05$ \\
Joint position noise & $\pm0.01$\,rad \\
Joint velocity noise & $\pm1.5$\,rad/s \\
Base push interval & $[1,3]$\,s \\
Base push velocity & $v_x,v_y:\pm0.5$\,m/s, $v_z:\pm0.2$\,m/s \\
Base push angular velocity & roll,pitch: $\pm0.52$\,rad/s, yaw: $\pm0.78$\,rad/s \\
Hand-force randomization & duration $[0.5,2.0]$\,s, capped at 20\,N \\
\bottomrule
\end{tabular}
\end{table}

\begin{table}[!h]
\centering
\caption{Reward terms used by all \yahmp{} variants.
Tracking rewards use $\exp(-e/\sigma^2)$; $F_{z,f}$ and $\mathbf{v}_{f,xy}$ are the vertical force and horizontal velocity of foot $f$.}
\label{tab:appendix_reward}
\scriptsize
\setlength{\tabcolsep}{2pt}
\begin{tabular}{@{}p{0.34\columnwidth}p{0.39\columnwidth}cc@{}}
\toprule
Term & Error~($e$) / cost & Weight & Std.~($\sigma$) \\
\midrule
Base pos. track. & $\|\mathbf{p}_b-\hat{\mathbf{p}}_b\|^2$ & 0.5 & 0.3 \\
Base ori. track. & $\mathrm{angle}(\hat{\mathbf{R}}_b^\top\mathbf{R}_b)^2$ & 0.5 & 0.4 \\
Body pos. track. & $\mathrm{mean}_k\|\mathbf{p}_k-\hat{\mathbf{p}}_k\|^2$ & 1.0 & 0.3 \\
Body ori. track. & $\mathrm{mean}_k\,\mathrm{angle}(\hat{\mathbf{R}}_k^\top\mathbf{R}_k)^2$ & 1.0 & 0.4 \\
Body lin. vel. track. & $\mathrm{mean}_k\|\mathbf{v}_k-\hat{\mathbf{v}}_k\|^2$ & 1.0 & 1.0 \\
Body ang. vel. track. & $\mathrm{mean}_k\|\boldsymbol{\omega}_k-\hat{\boldsymbol{\omega}}_k\|^2$ & 1.0 & 3.14 \\
Joint pos. track. & $\mathrm{mean}_j(q_j-\hat q_j)^2$ & 1.0 & 0.3 \\
Joint vel. track. & $\mathrm{mean}_j(\dot q_j-\dot{\hat q}_j)^2$ & 0.5 & 2.0 \\
Excess foot force & $\max(0,\|\mathbf{F}_z\|_2-300)$ & $-5\cdot 10^{-4}$ & -- \\
Foot slip & $\sum_{f:\, |F_{z,f}|>5}\sqrt{\|\mathbf{v}_{f,xy}\|}$ & $-0.1$ & -- \\
Action rate & $\|\mathbf{a}_t-\mathbf{a}_{t-1}\|^2$ & $-0.1$ & -- \\
Joint limits & soft-limit violation & $-10.0$ & -- \\
\bottomrule
\end{tabular}
\end{table}

\end{document}